\documentclass[11pt]{article}

\usepackage[final]{acl}

\usepackage{times}
\usepackage{latexsym}

\usepackage[T1]{fontenc}

\usepackage[utf8]{inputenc}

\usepackage{microtype}

\usepackage{inconsolata}

\usepackage{graphicx}

\usepackage{kotex}
\usepackage{amsmath}
\usepackage{amssymb}
\usepackage{booktabs}
\usepackage{graphicx}
\usepackage{fontawesome}
\usepackage{algorithm}
\usepackage{algpseudocode}

\usepackage[table]{xcolor}
\definecolor{myrow}{HTML}{EBEBFE}

\title{Generalizable Lifelong Model Editing via Preference Optimization}

\author{
  Dahyun Jung \quad \quad Suhyune Son \quad \quad Heuiseok Lim\thanks{Corresponding author.} \\
  Department of Computer Science and Engineering, Korea University \\
  \texttt{\{dhaabb55,ssh5131,limhseok\}@korea.ac.kr}
}

\begin{document}
\maketitle
\begin{abstract}

Knowledge editing enables rapid updates of specific factual knowledge in large language models (LLMs) without full retraining. However, more realistic scenarios call for a lifelong framework that handles continual updates rather than one-off modifications. In such settings, existing editing methods often overfit to target prompts, significantly degrading both the generalization of the edited knowledge and the model’s general capabilities. To address this issue, we propose GLIME (Generalizable Lifelong Model Editing), which combines knowledge editing with preference optimization over generation behavior. GLIME further incorporates replay-based editing and a gradient constraint to preserve previously edited knowledge. Experimental results show that GLIME significantly improves knowledge generalization in lifelong editing settings while maintaining both editing performance and general capabilities.\footnote{Our code is available at \url{https://github.com/ekgus9/GLIME}.}
\end{abstract}

\section{Introduction}
Large language models (LLMs) acquire extensive world knowledge and strong linguistic capabilities through pretraining~\cite{haviv2023understanding,openai2023gpt4,zhao2025surveylargelanguagemodels}, but their knowledge must be continually updated to reflect a changing world~\cite{decao2021editing,hartvigsen2023aging,shi2024continuallearninglargelanguage}. As a practical alternative to prohibitively expensive full retraining, knowledge editing has emerged as a promising approach for updating specific facts by directly modifying model parameters~\cite{mitchell2022memorybased}. However, in real-world scenarios, knowledge updates arrive not as isolated events but as an ongoing stream, requiring models to support lifelong knowledge editing under continual update requests~\cite{hartvigsen2023aging}.

In this lifelong setting, existing knowledge editing methods face two major limitations. First, they often achieve high editing success only on prompts that are lexically similar to the edited input, while generalizing poorly to novel contexts that require the edited knowledge~\cite{meng2023locating,zhong-etal-2023-mquake,yang-etal-2025-mirage}. This suggests that models tend to overfit to shallow lexical patterns surrounding specific triggers rather than internalizing the edited knowledge~\cite{li2024unveilingpitfallsknowledgeediting,ju2024investigating,thede2025wikibigeditunderstandinglimitslifelong}. Second, as edits accumulate, parameter updates gradually drift away from the original model distribution, thereby degrading the model's general capabilities~\cite{li-chu-2024-continually,gupta2024rebuildingromeresolving}. Such distributional shifts compound over time, eventually leading to a severe collapse in overall quality.

\begin{figure*}[hbt!]
\centering 
\includegraphics[width=0.95\textwidth]{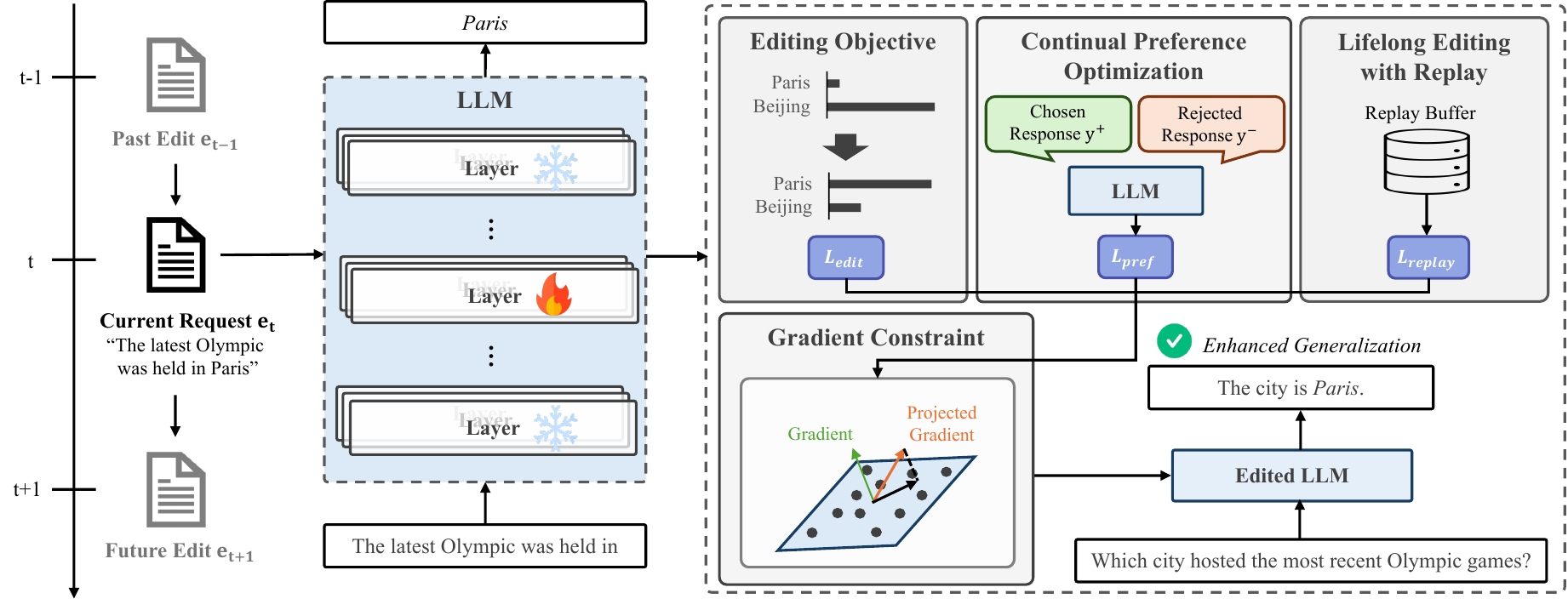}
\caption{Overall architecture of GLIME. \textit{Left}: The model is updated over a stream of continual knowledge edits using new knowledge injection, preference optimization for behavior-level regularization, and experience replay to prevent catastrophic forgetting. The gradient is further projected orthogonally to previous update directions to mitigate long-term collapse under continual edits. \textit{Bottom right}: The final edited model successfully answers paraphrased queries under autoregressive decoding, indicating improved generalization.
}
\label{fig:main} 
\end{figure*}

A primary underlying cause of this issue is that existing studies focus primarily on superficial objectives, such as edit success and locality on unrelated queries~\cite{wang2024wise, wang2023knowledge}. While these objectives are necessary, they do not ensure the generalization of edited knowledge or the preservation of the model's general capabilities. Consequently, methods that appear effective under restricted evaluation settings suffer severely from collapse due to overfitting as edits accumulate~\cite{yang-etal-2025-mirage}.

In this paper, we propose \textbf{GLIME (Generalizable Lifelong Model Editing)}, a framework designed to address the challenges of lifelong knowledge editing. As shown in Figure~\ref{fig:main}, GLIME treats continual editing not as a single-objective problem of merely increasing the probability of target tokens, but as a joint optimization problem that both injects new knowledge and preserves the model's generation capabilities. To this end, GLIME integrates preference optimization into the editing process to maintain the model’s general capabilities, thereby mitigating overfitting to target texts and facilitating generalized editing. 
Moreover, GLIME mitigates the catastrophic forgetting~\cite{gupta-etal-2024-model, luo2025empiricalstudycatastrophicforgetting} of past edits through two strategies. First, replay-based editing helps the model retain previously edited knowledge over time. Second, an orthogonal basis constraint prevents gradient updates for new edits from interfering with the feature subspace associated with past edits, thereby structurally protecting previously edited knowledge from being distorted.

To better assess the generalization of GLIME, we adopt a more realistic evaluation setting. Existing studies predominantly rely on teacher forcing-based evaluations, which tend to overestimate the model's generative capabilities by exposing the ground-truth targets~\cite{yang-etal-2025-mirage}. We therefore evaluate edited knowledge under autoregressive decoding, which more faithfully reflects its practical use. Experiments show that, compared with existing editing methods, GLIME significantly improves the generalization of edited knowledge, even with a large number of accumulated edits (e.g., 5k), while preserving editing performance and effectively mitigating model degradation.
Our main contributions are as follows:
\begin{itemize}
    \item We formulate lifelong knowledge editing as a problem of jointly integrating new knowledge, generalizing edited knowledge, and preserving the model's general capabilities.
    \item The proposed GLIME combines target editing with preference optimization to preserve general generation quality and reduce overfitting, while employing replay and gradient-space constraints to alleviate catastrophic forgetting.
    \item We introduce an autoregressive generation-based evaluation setting and show that GLIME improves the generalization and stability of lifelong editing.
\end{itemize}

\section{Related Work}

\paragraph{Model Editing}
Model editing updates specific knowledge in a pretrained language model by locally modifying its parameters, avoiding full retraining when correcting errors or incorporating new facts~\cite{zhu2020modifyingmemoriestransformermodels, decao2021editing}. Early approaches often rely on a small number of fine-tuning steps per edit request~\cite{zhu2020modifyingmemoriestransformermodels}. Later approaches improve efficiency and scalability through hypernetwork-based editors~\cite{decao2021editing, mitchell2022fast}, locate-then-edit methods~\cite{meng2023locating, meng2023massediting}, and modular editing with adapters or memory modules~\cite{hartvigsen2023aging}. ROME~\cite{meng2023locating} edits factual associations by identifying a causal multilayer perceptron (MLP) module and applying a rank-one update to write a new key--value pair. MEMIT~\cite{meng2023massediting} extends this idea to multi-layer updates, enabling large-scale simultaneous editing of many facts. 

\paragraph{Lifelong Editing}
In realistic settings, knowledge updates arrive not as isolated events but as an accumulating stream, requiring models to handle continual edit requests under lifelong editing~\cite{hartvigsen2023aging}. In this setting, continual edits not only cause forgetting of previously edited knowledge, but also gradually push model parameters away from the pretrained distribution, leading to model degradation in general capabilities~\cite{gupta2024rebuildingromeresolving}. To address this issue, GRACE~\cite{hartvigsen2023aging} introduces a discrete key--value adapter cache that preserves the original model weights while sequentially adding new knowledge. WISE~\cite{wang2024wise} proposes a dual-memory architecture with a main memory for pretrained knowledge and a side memory for edited knowledge, together with a router that selects between them. R-ROME~\cite{gupta2024rebuildingromeresolving} analyzes the instability of sequential ROME updates and introduces a more stable editing formulation to mitigate model collapse.

\paragraph{Continual Learning}
Lifelong knowledge editing is closely related to the problems of forgetting and interference in continual learning~\cite{mccloskey1989catastrophic}. To mitigate these issues, prior work has developed methods such as weight-importance regularization~\cite{Kirkpatrick_2017}, gradient projection and constraints~\cite{lopezpaz2022gradientepisodicmemorycontinual}, and experience replay~\cite{rolnick2019experience} to preserve performance on past data. More recent work extends these ideas to LLMs. GORP~\cite{wang-etal-2025-continual} projects continual fine-tuning updates into a unified low-rank gradient subspace across full-rank and low-rank parameters. \citet{abbes2025revisitingreplaygradientalignment} show that experience replay and gradient alignment improve stability in continual LLM pre-training.

Existing model editing methods either remain limited to simple knowledge memorization in constrained settings or fail to prevent model degradation under accumulated edits. In this study, we propose GLIME, which mitigates these issues through preference optimization and mitigates catastrophic forgetting by integrating an orthogonal subspace constraint with replay.

\section{Preliminaries}
\label{sec:preliminaries}

We denote a pretrained causal language model with parameters $\theta$ as a conditional distribution $p_\theta(\cdot \mid x)$. Given an input prompt $x$, the model generates a token sequence $y=(y_1,\dots,y_{|y|})$ autoregressively:
\begin{equation}
p_{\theta}(y \mid x)
= \prod_{i=1}^{|y|} p_{\theta}(y_i \mid x, y_{<i}).
\end{equation}

Model editing aims to locally update knowledge in a pretrained model $\theta$ by directly modifying its parameters with respect to a target fact. A single edit request $e$ consists of a query prompt $x$ and a target response $y^{\text{new}}$ that the edited model is expected to generate. In template-based settings, the query prompt is typically constructed as $x=\pi(s)$, where $\pi(\cdot)$ is a prompt template and $s$ is the subject. The goal of a single edit is to make the edited model assign higher probability to $y^{\text{new}}$ than to the original response $y$ for the query prompt $x$. The model is updated by an editing algorithm $\mathcal{E}$ as follows:
\begin{equation}
\theta' = \mathcal{E}(\theta, e).
\end{equation}

In realistic settings, knowledge update requests arrive not as isolated events but as a stream over time. We represent this process as a sequence of edit requests $\{e_t\}_{t=1}^{T}$, where each $e_t=(x_t, y_t^{\text{new}})$ specifies a distinct fact to be updated at time step $t$. Starting from an initial model $\theta_0$, the editing algorithm updates the parameters sequentially:
\begin{equation}
\theta_t = \mathcal{E}(\theta_{t-1}, e_t), \quad t=1,\dots,T.
\end{equation}

In lifelong editing, four criteria are typically considered when designing an editing algorithm~\cite{zhang2024comprehensive,fang2025alphaeditnullspaceconstrainedknowledge}. (i) \textit{Reliability}: the edited model correctly generates the target response $y_t^{\text{new}}$ for the requested prompt $x_t$. (ii) \textit{Generalization}: the edited knowledge is not tied only to the surface form of the original prompt, but can also be applied to paraphrases or queries in novel contexts. This is a key criterion for preventing edits from degenerating into local memorization around specific triggers. (iii) \textit{Locality}: the model's original behavior is preserved on queries $x$ unrelated to the edit. This reflects whether the update remains localized around the target fact while maintaining performance on unrelated inputs. (iv) \textit{General Capabilities}: the extent to which the model preserves its overall language quality and capabilities (e.g., language understanding, generation, reasoning, and instruction following) acquired during pretraining as edits accumulate. 

\section{Methodology}
\label{sec:method}

Our proposed GLIME (Generalizable Lifelong Model Editing) views lifelong knowledge editing not simply as a problem of correctly injecting new facts, but as one of mitigating overfitting to target texts, preventing model degradation under accumulated edits, and preserving past edits over time. Ultimately, GLIME aims to improve the generalization of edited knowledge. To this end, GLIME integrates three components within a unified editing loop: (i) continual preference optimization, (ii) replay-based continual editing, and (iii) a gradient-space constraint.

\subsection{Continual Preference Optimization}
\label{subsec:pref}

To maintain stable output quality under continual edits, GLIME incorporates preference optimization into the editing loop as a behavior-level regularizer.
Given an edit request $e=(x, y^{\text{new}})$, we first define the standard editing objective as follows:
\begin{equation}
\mathcal{L}_{\text{edit}}(\theta; e)         
=
-\log p_\theta\big(y^{\text{new}}\mid x\big).
\label{eq:edit_nll}
\end{equation}

However, this objective 
locally increases the probability of the target answer only for the given prompt, gradually pushing the model away from the original language distribution learned during pretraining as edits accumulate. As a result, the model's general generation ability may deteriorate during actual decoding.

GLIME therefore introduces preference optimization as a behavior-level constraint. Specifically, for a context $u$, we consider a preference pair consisting of a preferred response $y^+$ and a dispreferred response $y^-$. Here, $u$ is sampled from 
a preference dataset used 
for model alignment and is independent of the edit target prompt $x$. These preference pairs serve as a regularization signal that encourages the model to preserve its general preference distribution and generation quality throughout the continual editing process. Because $u$ is independent of the edit trigger, it also acts as an anchor that discourages the model from over-adapting to specific edit requests and distorting its output distribution.

More broadly, jointly optimizing these objectives encourages the model to find parameter updates that satisfy both factual editing and general behavioral consistency. By accommodating multiple objectives simultaneously, the model is guided toward shared effective structures rather than solutions that overfit to a single target, thereby improving generalization~\cite{ruder2017overviewmultitasklearningdeep}.

GLIME minimizes the following logistic preference loss~\cite{rafailov2024directpreferenceoptimizationlanguage, hong2024orpomonolithicpreferenceoptimization}:
\begin{equation}
\mathcal{L}_{\text{pref}}(\theta; u, y^+, y^-)
= -\log \sigma\!\left(\Delta_\theta(u)\right),
\label{eq:pref_1}
\end{equation}
\begin{equation}
\Delta_\theta(u)
=
\log p_\theta(y^+\mid u) - \log p_\theta(y^-\mid u),
\label{eq:pref_2}
\end{equation}
where $\sigma(\cdot)$ denotes the sigmoid function. This objective encourages the model to assign higher preference to $y^+$ than to $y^-$ under the context $u$.

\subsection{Replay for Lifelong Memory Retention}
\label{subsec:replay}

In the lifelong setting, optimizing only for the most recent edit $e_t$ degrades performance on previous edits $\{e_i\}_{i<t}$. To mitigate this issue, we maintain a replay buffer $\mathcal{B}_t$ that stores a subset of past edit requests and jointly train on them at the current step. Specifically, at time step $t$, we add an auxiliary editing loss over past requests sampled from $\mathcal{B}_t$:
\begin{equation}
\mathcal{L}_{\text{replay}}(\theta_t)
=
\mathbb{E}_{e\sim\mathcal{B}_t}\left[\mathcal{L}_{\text{edit}}(\theta_t; e)\right].
\label{eq:replay}
\end{equation}

The overall objective optimized by GLIME is then given by:
\begin{equation}
\mathcal{L}(\theta_t)
=
\mathcal{L}_{\text{edit}}(\theta_t; e_t)
+
\mathcal{L}_{\text{pref}}(\theta_t)
+
\lambda_{\text{rep}} \mathcal{L}_{\text{replay}}(\theta_t),
\end{equation}
where $\lambda_{\text{rep}}$ controls the contribution of the replay loss and is tuned to prevent the model from overfitting excessively to the current edit.
Following \citet{meng2023massediting,  fang2025alphaeditnullspaceconstrainedknowledge}, we restrict editing to the MLP layers that have been identified as playing a key role in factual prediction.

\subsection{Preventing Parameter Interference via Orthogonal Basis}
\label{subsec:long}

Unconstrained parameter optimization causes updates for newly injected knowledge to interfere with the weight subspace used by previous edits, leading to catastrophic forgetting of earlier knowledge. To mitigate this issue, inspired by \citet{farajtabar2019orthogonalgradientdescentcontinual,saha2021gradientprojectionmemorycontinual}, we introduce a gradient-space constraint that preserves update directions important for past edits and restricts new updates to the orthogonal complement of that subspace.

Let \(W \in \mathbb{R}^{m\times n}\) denote an editable weight matrix, and let \(G_t=\nabla_W \mathcal{L}(\theta_t)\in\mathbb{R}^{m\times n}\) be the gradient of the joint objective at step \(t\). To preserve information from previous edits, the model maintains an orthogonal basis matrix \(B_{t-1}\in\mathbb{R}^{n\times r}\), satisfying \(B_{t-1}^\top B_{t-1}=I\), which captures the right subspace of accumulated updates from earlier steps. The subspace spanned by \(B_{t-1}\), therefore, represents parameter directions that are important for retaining previously edited knowledge.

To minimize interference with past knowledge, GLIME projects the current gradient \(G_t\) onto the orthogonal complement of the existing basis:
\begin{equation}
\tilde{G}_t = G_t(I - B_{t-1}B_{t-1}^\top),
\label{eq:ogp}
\end{equation}
where \(\tilde{G}_t\) is orthogonal to the previous basis and thus satisfies \(\tilde{G}_t B_{t-1} = 0\). As a result, the current update suppresses parameter changes along directions important for past edits, while encouraging new knowledge to be learned in independent directions.

The basis \(B_t\) is updated online, and the detailed update procedure is described in Appendix~\ref{app:basis_update}. The final one-step parameter update in GLIME is:
\begin{equation}
W \leftarrow W - \eta \tilde G_t,
\label{eq:update}
\end{equation}
where \(\eta\) is the learning rate.

\section{Experiments}

\subsection{Evaluation Metrics}

We follow the standard evaluation framework of prior work~\cite{wang2024wise, wang2023knowledge}, which includes \textbf{Reliability}, \textbf{Generalization}, and \textbf{Locality}. However, conventional teacher forcing (TF)-based evaluation overestimates the usability of edited knowledge by exposing the ground-truth token during generation. To address this issue, we additionally adopt an autoregressive decoding (AD)-based protocol~\cite{yang-etal-2025-mirage}, in which the model generates the answer without access to the target token. This setting provides a more faithful assessment of how deeply the edited knowledge is integrated into the model's generation process. We further evaluate Portability and General Capabilities. More details on the evaluation metrics are provided in Appendix~\ref{sec:ap_metric}\footnote{
To mitigate the limitation of string-based matching in capturing semantic correctness, we provide an LLM-as-a-Judge in Appendix~\ref{subsec:llm_as_a_judge_analysis} and a qualitative analysis in Appendix~\ref{sec:qualitative_analysis}.}.

\paragraph{Portability}
Portability measures whether edited knowledge can be effectively applied to reasoning and downstream tasks, rather than being merely memorized as a direct answer. We consider two reasoning-based evaluations. (i) \textit{Multiple-Choice QA (MC)}: We evaluate whether the model can use edited knowledge to answer multiple-choice questions. Following \citet{su2024conflictbank}, each question targets an edited fact and provides four answer choices: the pre-edit answer, the post-edit answer, an unrelated answer, and an uncertain option. (ii) \textit{Multi-hop Reasoning QA (MR)}: We evaluate whether the model can use edited knowledge in multi-hop QA, following prior work~\cite{zhong-etal-2023-mquake, zhong2025mquakeremastered}, and report accuracy.

\paragraph{General Capabilities} 
An editing method should preserve the model's pretrained general capabilities after updates. To evaluate this property, we use five benchmarks: Winogrande~\cite{sakaguchi2019winograndeadversarialwinogradschema}, ARC~\cite{clark2018thinksolvedquestionanswering}, MathQA~\cite{amini-etal-2019-mathqa}, SQuADv2~\cite{rajpurkar-etal-2018-know}, and IFEval~\cite{zhou2023instructionfollowingevaluationlargelanguage}. All benchmarks are evaluated using the Language Model Evaluation Harness~\cite{eval-harness}\footnote{\url{https://github.com/EleutherAI/lm-evaluation-harness}}.

\begin{table*}[hbt!]
\centering
\small
\setlength{\tabcolsep}{3.2pt}
\resizebox{\textwidth}{!}{%
\begin{tabular}{l|cccccccc|c|cccccc|c}
\toprule
& \multicolumn{9}{c|}{\textbf{MQuAKE}} & \multicolumn{7}{c}{\textbf{ZSRE}} \\
\cmidrule(lr){2-10}\cmidrule(lr){11-17}
& \multicolumn{2}{c}{Reliability}
& \multicolumn{2}{c}{Generalization}
& \multicolumn{2}{c}{Locality}
& \multicolumn{2}{c|}{Portability}
& 
& \multicolumn{2}{c}{Reliability}
& \multicolumn{2}{c}{Generalization}
& \multicolumn{2}{c|}{Locality}
& \\
\cmidrule(lr){2-3}\cmidrule(lr){4-5}\cmidrule(lr){6-7}\cmidrule(lr){8-9}
\cmidrule(lr){11-12}\cmidrule(lr){13-14}\cmidrule(lr){15-16}
Method & TF & AD & TF & AD & TF & AD & MC & MR & AVG & TF & AD & TF & AD & TF & AD & AVG \\
\midrule

\multicolumn{17}{l}{\textbf{LLaMA-3.1-8B-Instruct}}\\ \midrule
FT-L      & 0.021 & 0.035 & 0.012 & 0.038 & 0.004 & 0.000 & 0.000 & 0.003 & 0.016 & 0.134 & 0.106 & 0.114 & 0.094 & 0.023 & 0.000 & 0.078 \\
R-ROME    & 0.026 & 0.074 & 0.009 & 0.087 & 0.005 & 0.010 & 0.000 & 0.008 & 0.030 & 0.033 & 0.004 & 0.028 & 0.087 & 0.000 & 0.000 & 0.025 \\
GRACE     & 0.310 & 0.035 & 0.203 & 0.046 & \textbf{0.571} & \textbf{0.714} & 0.010 & 0.060 & 0.270 & 0.378 & 0.021 & 0.314 & 0.014 & \textbf{0.388} & \textbf{0.203} & 0.220 \\
MEMIT     & 0.044 & 0.014 & 0.044 & 0.014 & 0.027 & 0.010 & 0.000 & 0.020 & 0.022 & 0.000 & 0.000 & 0.000 & 0.000 & 0.000 & 0.000 & 0.000 \\
WISE      & 0.479 & 0.193 & 0.433 & 0.147 & 0.328 & 0.121 & 0.000 & 0.028 & 0.243 & 0.379 & 0.073 & 0.366 & 0.055 & 0.322 & 0.012 & 0.201 \\
AlphaEdit & 0.912 & 0.604 & 0.497 & 0.514 & 0.324 & 0.209 & 0.434 & 0.079 & 0.499 & 0.882 & 0.538 & 0.820 & 0.451 & 0.384 & 0.187 & 0.544 \\ \midrule
\rowcolor{myrow}\textbf{GLIME}
          & \textbf{0.937} & \textbf{0.830} & \textbf{0.806} & \textbf{0.800} & 0.517 & 0.379 & \textbf{0.633} & \textbf{0.184} & \textbf{0.700} & \textbf{0.915} & \textbf{0.670} & \textbf{0.891} & \textbf{0.598} & 0.324 & 0.133 & \textbf{0.589} \\
\midrule

\multicolumn{17}{l}{\textbf{Qwen2.5-7B-Instruct}}\\ \midrule
FT-L      & 0.057 & 0.140 & 0.057 & 0.124 & 0.032 & 0.079 & 0.000 & 0.033 & 0.070 & 0.094 & 0.006 & 0.079 & 0.003 & 0.040 & 0.003 & 0.037 \\
R-ROME    & 0.000 & 0.001 & 0.000 & 0.001 & 0.000 & 0.001 & 0.000 & 0.006 & 0.000 & 0.226 & 0.054 & 0.217 & 0.050 & 0.055 & 0.003 & 0.101 \\
GRACE     & 0.349 & 0.050 & 0.179 & 0.055 & 0.373 & \textbf{0.684} & 0.009 & 0.079 & 0.243 & 0.414 & 0.024 & 0.322 & 0.020 & 0.381 & 0.100 & 0.210 \\
MEMIT     & 0.000 & 0.000 & 0.000 & 0.000 & 0.000 & 0.000 & 0.000 & 0.000 & 0.000 & 0.000 & 0.000 & 0.000 & 0.000 & 0.000 & 0.001 & 0.000 \\
WISE      & 0.512 & 0.274 & 0.466 & 0.202 & 0.260 & 0.124 & 0.008 & 0.063 & 0.264 & 0.463 & 0.231 & 0.443 & 0.197 & 0.239 & 0.040 & 0.269 \\
AlphaEdit & \textbf{0.986} & 0.201 & 0.369 & 0.389 & 0.361 & 0.513 & 0.192 & 0.081 & 0.430 & \textbf{0.984} & 0.354 & 0.870 & 0.277 & \textbf{0.383} & \textbf{0.113} & 0.497 \\ \midrule
\rowcolor{myrow}\textbf{GLIME}
          & 0.958 & \textbf{0.772} & \textbf{0.716} & \textbf{0.812} & \textbf{0.443} & 0.364 & \textbf{0.475} & \textbf{0.168} & \textbf{0.649} & 0.948 & \textbf{0.417} & \textbf{0.907} & \textbf{0.373} & 0.369 & 0.093 & \textbf{0.518} \\
\bottomrule
\end{tabular}%
}
\caption{Lifelong knowledge editing results.
We compare different editing methods on the MQuAKE and ZSRE benchmarks. For each dataset, we report Reliability, Generalization, and Locality. We additionally evaluate both teacher forcing (TF) and autoregressive decoding (AD), which reflects real generation settings. On MQuAKE, we further measure Portability using Multiple-Choice QA (MC) and Multi-hop Reasoning QA (MR), and report the overall average (AVG) across the presented metrics. Best results per model group are in \textbf{bold}.
}
\label{tab:main}
\end{table*}

\subsection{Datasets}
We evaluate editing accuracy on MQuAKE-Remastered~\cite{zhong2025mquakeremastered} and ZSRE~\cite{levy2017zero}. MQuAKE-Remastered is a corrected version of MQuAKE~\cite{zhong-etal-2023-mquake}, a dataset for multi-hop knowledge editing. From this benchmark, we use the CF-3k split. We perform edits using the original cloze prompts and use paraphrased question-style prompts to evaluate Generalization. For Locality, we select questions that do not overlap with those used for editing and construct a final test set of 1,000 samples. ZSRE is a context-free QA dataset designed for zero-shot relation extraction. After deduplicating the test set, we use 743 examples for evaluation. For preference optimization, we use OpenHermesPreferences~\cite{open_hermes_preferences}. More details on the datasets are provided in Appendix~\ref{sec:ap_dataset}.

\subsection{Implementation Details}
We conduct experiments on two instruction-tuned LLMs from different model families: LLaMA-3.1-8B-Instruct~\cite{dubey2024llama3herdmodels} and Qwen2.5-7B-Instruct~\cite{qwen2025qwen25technicalreport}. The corresponding checkpoints are `meta-llama/Llama-3.1-8B-Instruct' and `Qwen/Qwen2.5-7B-Instruct', both publicly available on Hugging Face\footnote{\url{https://huggingface.co/}}. Following EasyEdit~\cite{wang-etal-2024-easyedit}, we edit the MLP layers 4, 5, 6, 7, and 8. All experiments are conducted in a lifelong editing setting, where edits are applied one by one over the full dataset, and evaluation is performed after all edits have been completed. We train for 3 epochs with a learning rate of $5 \times 10^{-5}$ on a single RTX A6000 GPU. We set the basis size $k$ to 256, use 3 replay samples, and set the loss weight to 0.1. Additional experimental details are provided in Appendix~\ref{sec:ap_setup}, and hyperparameter sensitivity analyses are presented in Appendix~\ref{subsec:hyperparameter_analysis}.

\subsection{Baselines}
FT-L~\cite{zhu2020modifyingmemoriestransformermodels} injects target facts by fine-tuning a limited set of layers for each edit request. R-ROME~\cite{gupta2024rebuildingromeresolving} is a model editing method designed to reduce the model collapse observed in ROME~\cite{meng2023locating} and improve stability under continual edits. MEMIT~\cite{meng2023massediting} extends ROME to enable large-scale knowledge editing. GRACE~\cite{hartvigsen2023aging} trains a single layer, stores the resulting edited parameters in memory, and retrieves them when a relevant query is given. WISE~\cite{wang2024wise} introduces a dual-memory parameterization for lifelong editing, consisting of a main memory that stores pretrained knowledge and a side memory that stores edited knowledge. AlphaEdit~\cite{fang2025alphaeditnullspaceconstrainedknowledge} mitigates the interference caused by the perturbations used in locating-then-edit methods by constraining updates in the null space of existing knowledge. All implementations and baseline settings are based on EasyEdit~\cite{wang-etal-2024-easyedit}.

\subsection{Main Results}
\label{sec:main_results}

\begin{figure*}[hbt!]
\centering 
\includegraphics[width=\textwidth]{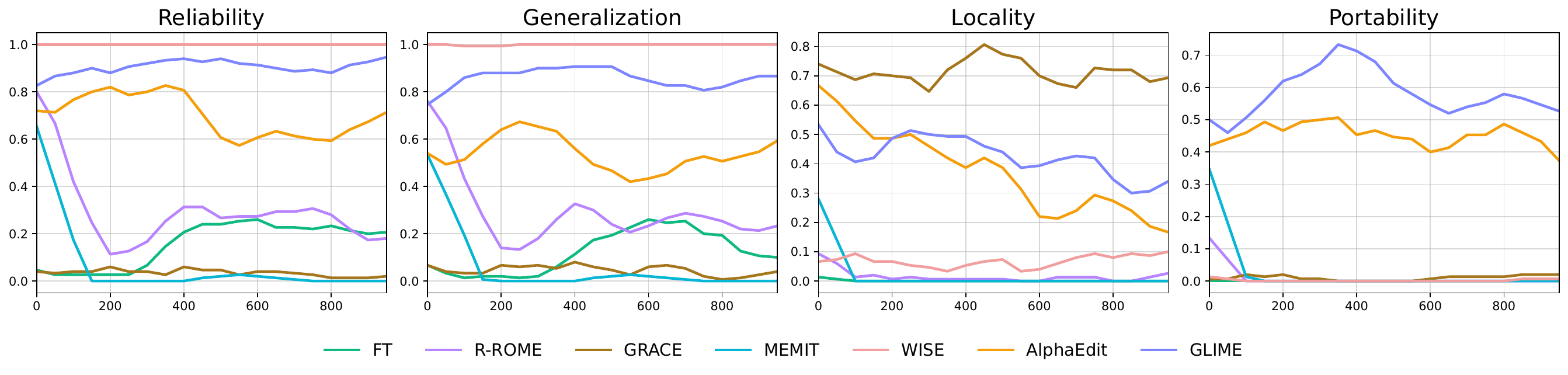}
\caption{Performance trends under lifelong editing as the number of edits increases. The figure compares performance degradation under accumulated edits and highlights differences in stability across methods.
}
\label{fig:seq} 
\end{figure*}

\begin{figure}[hbt!]
\centering 
\includegraphics[width=\linewidth]{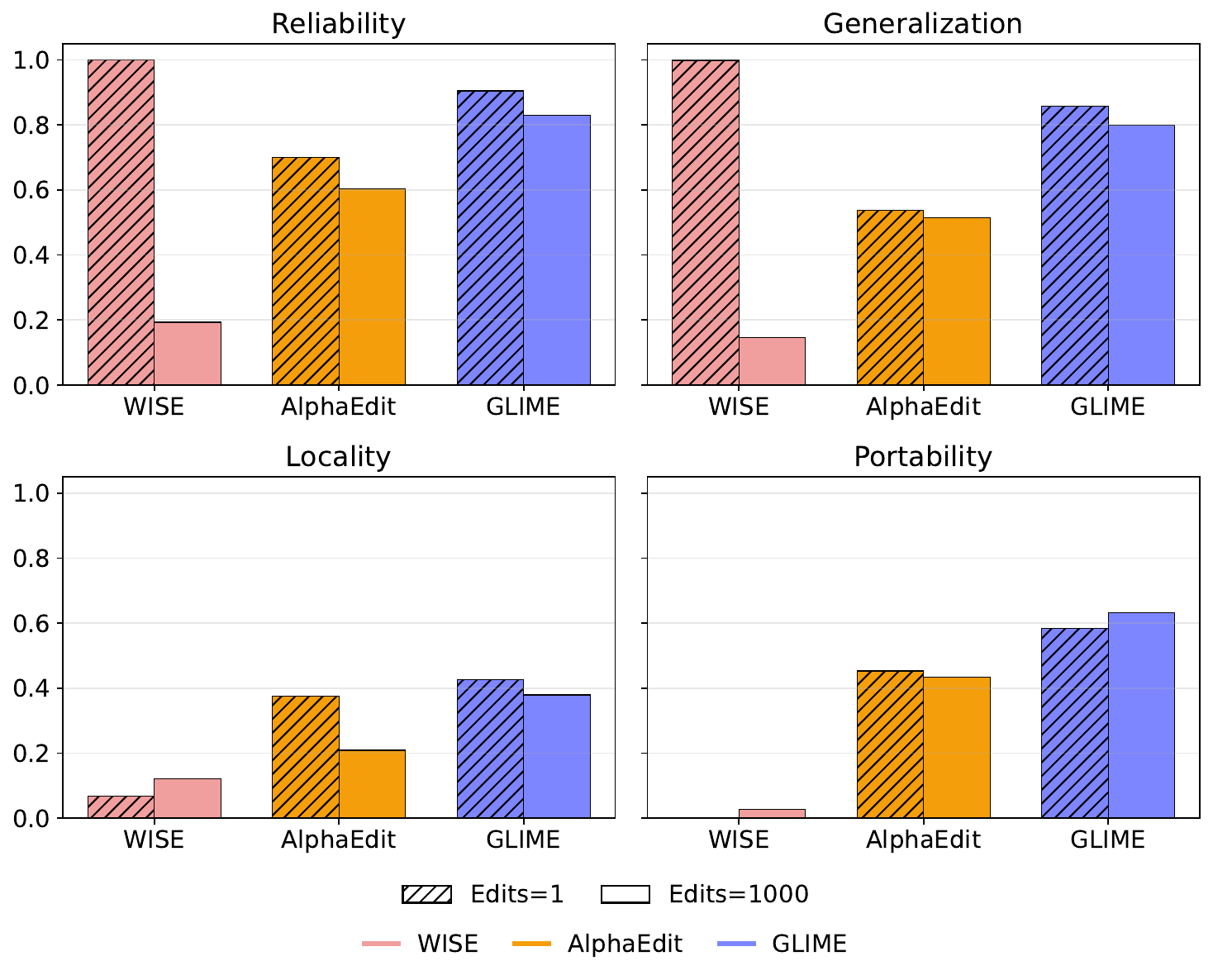}
\caption{Performance at different evaluation points. We report performance immediately after editing (Edits~=~1) and after accumulated edits (Edits~=~1000).
}
\label{fig:1000} 
\end{figure}

Table~\ref{tab:main} compares performance on MQuAKE-Remastered (MQuAKE) and ZSRE, reporting both TF and AD evaluation. Overall, GLIME achieves the most consistently strong performance across both backbones and both datasets, with particularly notable gains under AD.

\paragraph{Existing editing methods suffer substantial degradation under lifelong editing.}
FT-L, R-ROME, and MEMIT show generally low performance or rapid collapse under both TF and AD. WISE and GRACE achieve relatively strong results on some metrics, but fail to maintain a balanced trade-off between edit success and transfer performance, resulting in limited average performance. This suggests that existing methods struggle to maintain long-term stability and generalization under continual editing, as inter-edit interference and distribution drift accumulate over time.

\paragraph{TF evaluation tends to overestimate editing performance compared with AD.}
Across many baselines, high TF scores do not necessarily translate into strong AD performance. On MQuAKE with LLaMA-3.1-8B-Instruct, AlphaEdit achieves a high Reliability score of 0.912 under TF, but this drops sharply to 0.604 under AD. A similar pattern is observed on Qwen2.5-7B-Instruct, where performance appears overly tied to the edited prompt and degrades substantially under autoregressive generation. This gap highlights the difference between predicting the correct target token under TF and consistently applying the edited knowledge during actual generation. In the lifelong setting, this discrepancy becomes even more pronounced as edits accumulate, underscoring the importance of AD for evaluating the practical usability of edited knowledge.

\paragraph{GLIME shows the strongest robustness in knowledge generalization and reasoning-based use.}
GLIME maintains strong Reliability and Generalization on both MQuAKE and ZSRE across both backbones, while also achieving the largest gains on Portability measures that require reasoning. It further preserves Locality without substantial degradation relative to competing methods, alleviating the trade-off between edit success and non-target preservation. These results show that GLIME goes beyond producing the correct answer for the original edit prompt: it enables the model to apply edited knowledge more reliably during generation while mitigating overall performance degradation under accumulated edits.

\section{Analysis}

Unless otherwise stated, all analysis experiments are conducted on MQuAKE using LLaMA-3.1-8B-Instruct and evaluated under AD. For Portability, we report MC scores.

\paragraph{Stability under Lifelong Editing}
Figure~\ref{fig:seq} shows the performance trajectories of different methods as the number of edits increases. Evaluation is performed immediately after each edit to identify the point at which model collapse emerges under accumulated updates. FT-L and GRACE exhibit poor generalization of edited knowledge even at an early stage, suggesting they focus too narrowly on the edit prompt itself. R-ROME and MEMIT show strong edit generalization at first, but undergo sharp model collapse as edits accumulate. WISE produces the edited knowledge accurately immediately after editing, but its low Portability indicates limited ability to use that knowledge in downstream reasoning. AlphaEdit remains relatively strong and robust in edit generalization under accumulated edits, but shows a gradual decline in Locality over time. In contrast, GLIME exhibits smaller performance degradation throughout the entire editing sequence and maintains smoother performance curves, indicating more stable knowledge updating even in long edit streams.

\begin{table*}[t]
\centering
\small
\setlength{\tabcolsep}{3.2pt}
\resizebox{0.7\linewidth}{!}{%
\begin{tabular}{ccc|ccccc}
\toprule
CPO & Replay & GC
& Reliability & Generalization & Locality
& Portability$_{\mathrm{MC}}$ & Portability$_{\mathrm{MR}}$ \\
\midrule

\rowcolor{myrow}
\checkmark & \checkmark & \checkmark
& \textbf{0.830}
& \textbf{0.800}
& 0.379
& \textbf{0.633}
& \textbf{0.184} \\ \midrule

& \checkmark & \checkmark
& 0.826 {\textcolor{red}{\scriptsize(-0.004)}}
& 0.745 {\textcolor{red}{\scriptsize(-0.055)}}
& 0.187 {\textcolor{red}{\scriptsize(-0.192)}}
& 0.014 {\textcolor{red}{\scriptsize(-0.619)}}
& 0.112 {\textcolor{red}{\scriptsize(-0.072)}} \\

\checkmark & & \checkmark
& 0.657 {\textcolor{red}{\scriptsize(-0.173)}}
& 0.633 {\textcolor{red}{\scriptsize(-0.167)}}
& 0.326 {\textcolor{red}{\scriptsize(-0.053)}}
& 0.531 {\textcolor{red}{\scriptsize(-0.102)}}
& 0.096 {\textcolor{red}{\scriptsize(-0.088)}} \\

\checkmark & \checkmark &
& 0.800 {\textcolor{red}{\scriptsize(-0.030)}}
& 0.786 {\textcolor{red}{\scriptsize(-0.014)}}
& 0.385 {\textcolor{red}{\scriptsize(+0.006)}}
& 0.412 {\textcolor{red}{\scriptsize(-0.221)}}
& 0.098 {\textcolor{red}{\scriptsize(-0.086)}} \\

\checkmark & &
& 0.597 {\textcolor{red}{\scriptsize(-0.233)}}
& 0.601 {\textcolor{red}{\scriptsize(-0.199)}}
& \textbf{0.391} {\textcolor{red}{\scriptsize(+0.012)}}
& 0.530 {\textcolor{red}{\scriptsize(-0.103)}}
& 0.168 {\textcolor{red}{\scriptsize(-0.016)}} \\

& \checkmark &
& 0.624 {\textcolor{red}{\scriptsize(-0.206)}}
& 0.615 {\textcolor{red}{\scriptsize(-0.185)}}
& 0.137 {\textcolor{red}{\scriptsize(-0.242)}}
& 0.063 {\textcolor{red}{\scriptsize(-0.570)}}
& 0.109 {\textcolor{red}{\scriptsize(-0.075)}} \\

& & \checkmark
& 0.528 {\textcolor{red}{\scriptsize(-0.302)}}
& 0.533 {\textcolor{red}{\scriptsize(-0.267)}}
& 0.259 {\textcolor{red}{\scriptsize(-0.120)}}
& 0.313 {\textcolor{red}{\scriptsize(-0.320)}}
& 0.120 {\textcolor{red}{\scriptsize(-0.064)}} \\

& &
& 0.158 {\textcolor{red}{\scriptsize(-0.672)}}
& 0.120 {\textcolor{red}{\scriptsize(-0.680)}}
& 0.141 {\textcolor{red}{\scriptsize(-0.238)}}
& 0.098 {\textcolor{red}{\scriptsize(-0.535)}}
& 0.087 {\textcolor{red}{\scriptsize(-0.097)}} \\

\bottomrule
\end{tabular}
}
\caption{
Ablation results for GLIME across all combinations of its core components.
Values in parentheses indicate performance changes relative to the full GLIME.
CPO denotes Continual Preference Optimization, Replay denotes the replay loss,
and GC denotes the gradient-space constraint.
}
\label{tab:ablation}
\end{table*}

Figure~\ref{fig:1000} compares performance under different evaluation points. Comparing performance immediately after editing with that after many accumulated edits shows that some methods perform well in the single-edit setting but deteriorate substantially as edits accumulate. WISE achieves very high Reliability and Generalization at Edits~=~1, but both metrics drop sharply at Edits~=~1000, suggesting that continual updates induce substantial model drift. By contrast, GLIME achieves stronger editing performance than competing methods already at Edits~=~1, while also preserving knowledge transfer and model stability under accumulated edits. These results show that GLIME is designed not only for strong one-shot editing performance, but also for robust lifelong editing.

\paragraph{Ablation Results}
Table~\ref{tab:ablation} summarizes the contribution of each core component of GLIME across all possible combinations. Removing CPO substantially weakens Generalization and Portability. Notably, using Replay and GC without CPO maintains relatively high Reliability and Generalization (0.826 and 0.745, respectively), but results in substantially lower Portability, particularly on MC (0.014). This indicates that these continual learning components mainly contribute to retaining previously edited knowledge, but are insufficient to improve its portability. Removing Replay notably reduces Reliability, indicating weaker retention of previously edited knowledge. Removing GC particularly hurts Portability, suggesting that GC helps prevent interference across edits and preserve the usable integration of edited knowledge. Overall, Replay and GC primarily support the retention of past edits in the lifelong setting, whereas CPO plays a key role in improving the generalization and portability of edited knowledge.

\begin{figure}[t]
\centering 
\includegraphics[width=\linewidth]{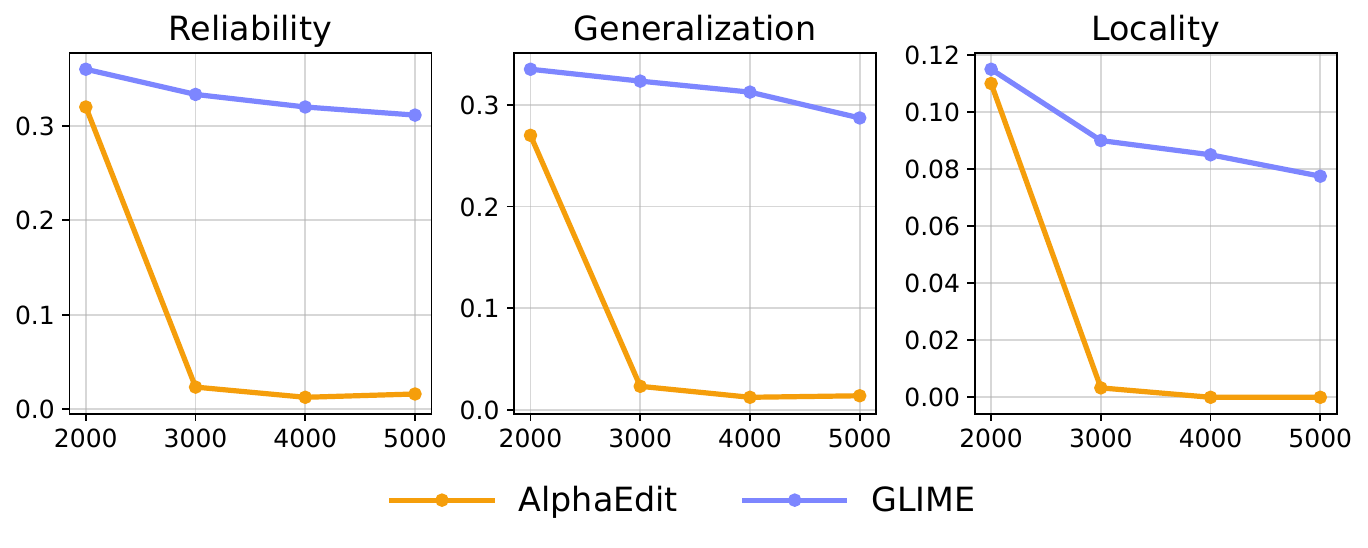}
\caption{Performance under up to 5,000 edits.}
\label{fig:5000} 
\end{figure}

\begin{table}[t]
\centering
\small
\setlength{\tabcolsep}{3.2pt}
\resizebox{\linewidth}{!}{%
\begin{tabular}{l|ccccc}
\toprule
Method & Winogrande & ARC & MathQA & Squadv2 & IFEval \\
\midrule
Base      & 0.748 & 0.802 & 0.390 & 0.503 & 0.532 \\ \midrule
FT-L      & 0.248 & 0.270 & 0.188 & 0.002 & 0.171  \\
R-ROME    & 0.514 & 0.234 & 0.226 & 0.000 & 0.229 \\
GRACE     & 0.748 & 0.802 & 0.390 & 0.503 & 0.532 \\
MEMIT     & 0.500 & 0.244 & 0.178 & 0.000 & 0.184 \\
WISE      & 0.738 & 0.778 & 0.394 & 0.129 & 0.209 \\
AlphaEdit & 0.716 & 0.772 & 0.404 & 0.654 & 0.481 \\ \midrule
\rowcolor{myrow}\textbf{GLIME}      & 0.716 & 0.772 & 0.412 & 0.596 & 0.486 \\ 
\bottomrule
\end{tabular}
}
\caption{General capability results. We evaluate each editing method on representative benchmarks unrelated to the edited knowledge. \textit{Base} denotes the performance of the model before editing, while the other results are measured on the final model after accumulated edits.}
\label{tab:results}
\end{table}

\begin{table}[t]
\centering
\small
\setlength{\tabcolsep}{3.2pt}
\resizebox{\linewidth}{!}{%
\begin{tabular}{l|ccccccc}
\toprule
Method & FT-L                                                    & R-ROME                                                  & GRACE                                                   & MEMIT                                                   & WISE                                                    & AlphaEdit                                               & \textbf{GLIME}                                           \\ \midrule
Seconds & \multicolumn{1}{r}{\cellcolor[HTML]{FFFFFF}2.574} & \multicolumn{1}{r}{\cellcolor[HTML]{F2F2FE}8.827} & \multicolumn{1}{r}{\cellcolor[HTML]{F8F8FF}5.975} & \multicolumn{1}{r}{\cellcolor[HTML]{C7C7FA}27.826} & \multicolumn{1}{r}{\cellcolor[HTML]{F5F5FF}7.494} & \multicolumn{1}{r}{\cellcolor[HTML]{E1E1FD}16.431} & \multicolumn{1}{r}{\cellcolor[HTML]{F1F1FE}8.936} \\ \bottomrule
\end{tabular}
}
\caption{Computational cost comparison. We report the time required for a single edit for each method.
}
\label{tab:time}
\end{table}

\paragraph{Scaling with Longer Sequences of Edits}
A lifelong model editing should maintain stable performance as the number of edits increases. To evaluate scalability, we extend the total number of edits to 5,000 and conduct editing on the ZSRE training set. Figure~\ref{fig:5000} compares GLIME with AlphaEdit, showing that while AlphaEdit suffers a sharp overall performance drop after around 3,000 edits, GLIME remains substantially more stable and scalable throughout the full editing sequence.

\paragraph{Preservation of General Capabilities}
Table~\ref{tab:results} compares performance on downstream benchmarks unrelated to the edited knowledge after editing. FT-L, R-ROME, and MEMIT show substantial degradation on most tasks, suggesting that continual parameter updates accumulate drift from the pretrained distribution and eventually impair the model's overall language and reasoning abilities. GLIME maintains general performance comparable to AlphaEdit and remains stable across diverse tasks, with only small drops relative to the Base model. These results indicate that GLIME is not overly biased toward maximizing edit success alone, but instead effectively mitigates degradation by preserving the model's pretrained general capabilities.

\paragraph{Time and Efficiency Analysis}
Table~\ref{tab:time} reports the time required for a single edit for each method. In lifelong settings, editing must be efficient enough to support rapid updates. GLIME remains practical while achieving competitive efficiency.

\section{Conclusion}
In this paper, we introduced GLIME, a framework for lifelong knowledge editing in LLMs that addresses superficial overfitting, degradation of general capabilities, and catastrophic forgetting. GLIME combines knowledge injection with preference optimization for preserving generation capabilities, encouraging edited knowledge to generalize beyond specific triggers rather than being shallowly memorized. In addition, it incorporates replay and a gradient-space constraint to mitigate forgetting under accumulated edits. Experiments showed that GLIME degrades more gracefully than strong editing baselines over long edit sequences and enables edited knowledge to transfer more reliably across diverse contexts.

\section*{Limitations}

GLIME improves generalization and long-term stability under lifelong model editing, but it has several limitations. First, it incurs additional computational and memory overhead by maintaining a replay buffer and a low-rank orthogonal basis of past gradients. Although we validate its effectiveness up to 5,000 edits, real-world deployment may involve much longer edit streams, motivating more efficient replay and basis management. We adopt simple and relatively memory-efficient mechanisms, but exploring more scalable alternatives remains an important direction for future work. Second, its scalability across model sizes and architectures remains unclear. Our experiments are limited to 7B--8B models and specific MLP layers, and future work should examine whether GLIME remains effective in much larger models or architectures such as Mixture-of-Experts (MoE). Finally, our evaluation focuses on mutually unrelated factual knowledge. While MQuAKE and ZSRE allow us to assess explicit factual updates and their transferability, real-world LLM deployment involves more diverse types of knowledge, and the effectiveness of GLIME in such settings remains to be studied.

\section*{Acknowledgments}
This research was supported by Basic Science Research Program through the National Research Foundation of Korea(NRF) funded by the Ministry of Education(NRF-2021R1A6A1A03045425).
This work was supported by Institute for Information \& communications Technology Promotion(IITP) grant funded by the Korea government(MSIT) 
(RS-2024-00398115, Research on the reliability and coherence of outcomes produced by Generative AI).
This work was supported by Institute for Information \& communications Technology Planning \& Evaluation(IITP) grant funded by the Korea government(MSIT) (No. RS-2022-II220369, (Part 4) Development of AI Technology to support Expert Decision-making that can Explain the Reasons/Grounds for Judgment Results based on Expert Knowledge).
This work was supported by the Commercialization Promotion Agency for R\&D Outcomes(COMPA) grant funded by the Korea government(Ministry of Science and ICT)(2710096072).


\bibliography{custom}

\clearpage

\appendix

\section{Online Update of Basis}
\label{app:basis_update}

This section details how GLIME updates the orthogonal basis memory \(B_t\) after each edit step.
The goal is to preserve parameter directions that were important for previous edits while allowing the current edit to use only directions in the orthogonal complement.

After the parameter update, we augment the memory with the new right-subspace used by the current edit.
A direct SVD of \(G_t\) or \(\tilde{G}_t\) at every step would be expensive for large layers.
Instead, we extract a compact low-rank approximation of the right subspace using a randomized sketch~\cite{halko2010findingstructurerandomnessprobabilistic}.

Specifically, let \(k\) denote the sketch dimension, and sample a random Gaussian matrix
\(\Omega_t \in \mathbb{R}^{m \times k}\).
We then form the sketch:
\begin{equation}
Y_t = \tilde{G}_t^{\top}\Omega_t \in \mathbb{R}^{n \times k}.
\label{eq:sketch}
\end{equation}

Since \(Y_t\) is obtained by multiplying \(\tilde{G}_t^{\top}\) with a random test matrix, the column space of \(Y_t\) provides a low-dimensional approximation to the dominant right subspace of the current projected gradient.

Next, we compute a thin-QR factorization:
\begin{equation}
Y_t = Q_t S_t,
\label{eq:local_qr}
\end{equation}
where \(Q_t \in \mathbb{R}^{n \times \hat r_t}\) has orthonormal columns, \(S_t \in \mathbb{R}^{\hat r_t \times k}\), and \(\hat r_t \le k\) is the numerical rank of the sketch.
We use \(Q_t\) as the compact basis of the new right-subspace induced by the current edit.

To update the global memory, we concatenate the previous basis and the newly extracted basis:
\begin{equation}
\bar{B}_t = \left[\, B_{t-1}\;\; Q_t \,\right].
\label{eq:concat_basis}
\end{equation}

Although \(\tilde{G}_t\) is orthogonal to \(B_{t-1}\) in exact arithmetic, a re-orthogonalization step is applied for numerical stability.
Concretely, we perform another thin-QR factorization,
\begin{equation}
\bar{B}_t = \hat{Q}_t \hat{S}_t,
\label{eq:global_qr}
\end{equation}
and set
\begin{equation}
B_t = \hat{Q}_t.
\end{equation}

By construction, \(B_t^\top B_t = I\), and \(B_t\) spans both the previously protected right-subspace and the new right directions introduced at step \(t\).

We do not claim that this approach is the only valid choice. We mitigate catastrophic forgetting using simple, relatively memory-efficient replay and gradient-constrained methods. Exploring more sophisticated and efficient approaches would be an important direction for future work.

\section{Evaluation Metric Details}
\label{sec:ap_metric}

We adopt four standard knowledge editing metrics, along with a General Capabilities metric, to provide a comprehensive evaluation of lifelong knowledge editing. In particular, to address the limitations of teacher forcing (TF)-based evaluation commonly used in prior work and to better reflect real deployment settings, we treat autoregressive decoding (AD)-based evaluation as a core metric~\cite{yang-etal-2025-mirage}.

\subsection{Difference between TF and AD}
\label{subsec:tf_vs_ad}
Before defining the evaluation metrics, we first clarify the distinction between TF and AD.

\paragraph{Teacher Forcing}
In TF, when predicting a target sequence token by token, the model predicts the next token $\hat{y}_j$ under the assumption that all previous tokens are given as the ground-truth prefix $y_{<j}^*$. Formally, given a prompt $x$ and a target sequence $y^* = (y_1^*, y_2^*, \dots, y_L^*)$, the predicted token at each step $j$ is defined as:
\begin{equation}
\hat{y}_j = \arg\max_{v \in \mathcal{V}} p_\theta(v \mid x, y_{<j}^*).
\end{equation}

Because the input at each step is corrected with the ground-truth token even when the model makes an incorrect prediction at the previous step, distortions in the generation distribution do not accumulate. As a result, TF tends to overestimate the model's actual decoding ability.

\paragraph{Autoregressive Decoding}
In AD, the model reuses its own previously generated tokens as input at subsequent steps. Given only a prompt $x$, the generation function $\text{Decode}(\theta, x)$ produces tokens sequentially. Under greedy decoding, the predicted token at step $j$ is defined as:
\begin{equation}
\hat{y}_j = \arg\max_{v \in \mathcal{V}} p_\theta(v \mid x, \hat{y}_{<j}).
\end{equation}

Unlike TF, AD conditions on the model's own predictions rather than the ground-truth prefix. As a result, a single incorrect token diverts the entire subsequent generation trajectory through $\hat{y}_{<j}$, making AD much more sensitive to exposure bias. This is one reason why existing editing methods often perform well under TF but degrade substantially under AD after aggressive parameter updates. GLIME addresses this issue through a behavior-level constraint based on continual preference optimization.

\subsection{Reliability}
Reliability measures whether the model can correctly generate the new target response $y_t^{\text{new}}$ for a given prompt $x_t$.
\begin{itemize}
    \item \textbf{TF:} Let $L$ be the length of the target response $y_t^{\text{new}}$. For each target position $j \in \{1,\dots,L\}$, we condition on the ground-truth prefix $y_{t,<j}^{\text{new}}$ and compare the model's predicted token with the target token $y_{t,j}^{\text{new}}$. Reliability under TF is defined as the average token-level accuracy over the $L$ target positions.
    \item \textbf{AD:} Given only the prompt $x_t$, we generate an output sequence $\hat{y}_t = \text{Decode}(\theta_t, x_t)$ and evaluate accuracy based on whether the target answer text is exactly contained in the generated output.
\end{itemize}

\begin{table*}[h]
\centering
\small
\renewcommand{\arraystretch}{1.3}
\begin{tabular}{p{2cm}p{3cm}p{6cm}p{3cm}}
\toprule
\textbf{Dataset} & \textbf{Task Category} & \textbf{Input Prompt ($x$)} & \textbf{Target Output ($y$)} \\
\midrule
\textbf{MQuAKE}
& \textbf{Base Edit} (Rel.) & jazz was created in the country of & Indonesia \\
& \textbf{Generalization} & Which country was jazz created in? & Indonesia \\
& \textbf{Locality} & What type of music does Hamid Drake play? & jazz \\
& \textbf{Portability (MC)} & jazz was created in the country of \newline \textit{Options: (A) Indonesia, (B) uncertain, (C) United States of America, (D) jazz} & (A) \\
& \textbf{Portability (MR)} & In which country was the music genre played by Hamid Drake created? & Indonesia \\
\midrule
\textbf{ZSRE}
& \textbf{Base Edit} (Rel.) & What programming language was used to write OpenCV? & Java \\
& \textbf{Generalization} & What is the language of OpenCV? & Java \\
& \textbf{Locality} & nq question: who was the head of the spanish inquisition & Grand Inquisitor \\
\bottomrule
\end{tabular}
\caption{Example knowledge editing instances from the MQuAKE and ZSRE datasets.}
\label{tab:edit_dataset_examples}
\end{table*}

\begin{table*}[h]
\centering
\small
\renewcommand{\arraystretch}{1.2}
\begin{tabular}{l| ccc| ccc}
\toprule
& \multicolumn{3}{c|}{\textbf{MQuAKE}} & \multicolumn{3}{c}{\textbf{ZSRE}} \\
\cmidrule(lr){2-4} \cmidrule(lr){5-7}
& \textbf{Reliability} & \textbf{Generalization} & \textbf{Locality} & \textbf{Reliability} & \textbf{Generalization} & \textbf{Locality} \\
\midrule
\textbf{Target True} & 0.646 & 0.418 & 0.571 & 0.438 & 0.428 & 0.388 \\
\textbf{Target New} & 0.200 & 0.201 & 0.571 & 0.331 & 0.326 & 0.388 \\
\bottomrule
\end{tabular}
\caption{Pre-edit performance of LLaMA-3.1-8B-Instruct before editing.}
\label{tab:base_performance}
\end{table*}

\subsection{Generalization}
Generalization measures whether the edited fact extends beyond surface-level memorization of a specific trigger and can be successfully elicited from a set of $K$ semantically equivalent paraphrased queries, $\mathcal{X}_t^{\text{gen}} = \{x_{t,1}^{\text{gen}}, \dots, x_{t,K}^{\text{gen}}\}$, under both TF and AD evaluation.

\subsection{Locality}
Locality measures whether the model preserves its original world knowledge on a set of prompts unrelated to the target fact, $\mathcal{X}_t^{\text{loc}}$, under both TF and AD evaluation.

\subsection{Portability}
Portability evaluates whether the injected knowledge can be used in downstream tasks or reasoning.
\begin{itemize}
    \item \textbf{Multiple-Choice QA (MC):} We present the model with a four-way multiple-choice question $x^{\text{MC}}$ and let it generate the answer directly. We then check whether the generated output, $\text{Decode}(\theta_t, x^{\text{MC}})$, explicitly identifies the target answer option $y^{\text{new}}$.
    \item \textbf{Multi-hop Reasoning QA (MR):} We provide a question $x^{\text{MR}}$ in which the edited knowledge must be used as an intermediate step in a reasoning chain, and evaluate accuracy based on whether the final reasoning answer is correctly contained in $\text{Decode}(\theta_t, x^{\text{MR}})$.
\end{itemize}

\subsection{General Capabilities}
\label{subsec:general_capability}
As knowledge updates accumulate, biased parameter updates may erode the general capabilities that a language model possessed before editing. To assess this effect from multiple perspectives, we evaluate the edited models on five representative NLP benchmarks using the Language Model Evaluation Harness~\cite{eval-harness}.

\paragraph{Winogrande~\cite{sakaguchi2019winograndeadversarialwinogradschema}}
Winogrande is a large-scale commonsense reasoning benchmark and an extension of the Winograd Schema Challenge (WSC) for pronoun resolution. We use it to assess whether the model, after knowledge edits, retains the physical and social commonsense needed to correctly infer sentence meaning from context. Accuracy is computed as the fraction of correct options that receive the highest length-normalized log-probability among the candidate choices.

\paragraph{ARC~\cite{clark2018thinksolvedquestionanswering}}
ARC consists of grade-school science exam questions, including a challenging subset that cannot be solved by simple pattern matching or surface-level retrieval alone. We use this benchmark to evaluate whether continual forced parameter updates damage the model's underlying scientific reasoning ability and world knowledge.

\begin{table*}[h]
\centering
\small
\renewcommand{\arraystretch}{1.4}
\begin{tabular}{p{2.5cm} p{12.5cm}}
\toprule
\textbf{Role} & \textbf{Content} \\
\midrule
\textbf{User Query ($u$)} & What's the most reliable medical treatment for premature graying? \\
\midrule
\textbf{Chosen ($y^+$)} & At present, there is no scientifically proven, consistently reliable medical treatment for premature graying, as the causes are often unknown and multifactorial. Graying hair is a natural part of the aging process when the pigment-producing cells in hair follicles slowly reduce their production of melanin, which gives hair its natural color. However, in some cases, premature hair graying can be a result of genetics, stress, certain medical conditions, or other lifestyle factors. \newline \newline If you are concerned about premature graying, it's essential to consult with a healthcare professional, who can help determine the underlying causes and potential treatment options based on your individual case and your specific needs. They may recommend addressing any imbalances in your healthcare plan, improving overall well-being, and monitoring any related medical conditions. \\
\midrule
\textbf{Rejected ($y^-$)} & There is currently no definitive medical treatment to stop or reverse premature graying. The graying of hair is primarily influenced by genetics. However, if there is an underlying medical condition causing premature graying, treating that condition may help slow down the process. It's always best to consult with a healthcare provider or a dermatologist for personalized advice. \newline \newline In some cases, premature graying can be a sign of vitamin B12 deficiency, so addressing this deficiency might help. Certain hair care products and hair dyes can also help to mask the appearance of gray hair. \newline \newline It's important to note that while there are many supplements and home remedies advertised to prevent or reverse graying, there is little scientific evidence to support these claims. \\
\bottomrule
\end{tabular}
\caption{Example of preference pairs from the OpenHermesPreferences dataset.}
\label{tab:pref_dataset_examples}
\end{table*}

\begin{table*}[t]
\centering
\small
\resizebox{\textwidth}{!}{%
\begin{tabular}{llll}
\toprule
\textbf{Phase} & \textbf{Template Type} & \textbf{Input Prompt ($x$)} & \textbf{Target Output ($y$)} \\
\midrule
Editing
& Context-free cloze
& jazz was created in the country of
& Indonesia \\
\midrule
Evaluation
& QA prompt with official chat template
& Question: Which country was jazz created in? Answer:
& Indonesia \\
\bottomrule
\end{tabular}
}
\caption{Example prompt templates used in the editing and evaluation phases.}
\label{tab:prompt_templates}
\end{table*}

\paragraph{MathQA~\cite{amini-etal-2019-mathqa}}
MathQA is a benchmark for solving complex multi-step math word problems. Although knowledge editing primarily modifies factual associations, severe parameter drift may also harm the structural reasoning pathways required for mathematical problem solving. This benchmark measures the extent to which such degradation affects mathematical reasoning ability.

\paragraph{SQuAD v2~\cite{rajpurkar-etal-2018-know}}
SQuAD v2 evaluates machine reading comprehension on Wikipedia passages. We report the F1 score between the model prediction and the reference answer. Importantly, SQuAD v2 includes unanswerable questions for which no correct answer is present in the provided context. This makes it particularly useful for assessing whether knowledge editing increases hallucination or causes the model to answer too aggressively when it should abstain.

\paragraph{IFEval~\cite{zhou2023instructionfollowingevaluationlargelanguage}}
IFEval measures how strictly a model follows instructions that impose specific formats or constraints. Given a set of $N$ constraints $\mathcal{C} = \{c_1, \dots, c_N\}$ specified in the prompt (e.g., starting with a capital letter or avoiding a particular word), we compute the proportion of constraints satisfied by the generated text using strict accuracy. This benchmark verifies whether an instruction-tuned model retains its instruction-following ability after knowledge editing and can continue to respond safely and appropriately to user requests.

\section{Dataset Details}
\label{sec:ap_dataset}
We evaluate knowledge editing performance using the MQuAKE-Remastered~\cite{zhong2025mquakeremastered} and ZSRE~\cite{levy2017zero} datasets, and use OpenHermesPreferences~\cite{open_hermes_preferences} as anchor data for continual preference optimization to suppress behavior-level distribution drift. The detailed composition of each dataset and example instances used in the actual benchmark evaluation are described below.

\subsection{Knowledge Editing Evaluation Datasets}
To compare the editing and retention capabilities of GLIME with those of existing baselines, we adopt MQuAKE-Remastered and ZSRE, two widely used benchmarks for knowledge editing.

\paragraph{MQuAKE-Remastered}
MQuAKE~\cite{zhong-etal-2023-mquake} is originally designed to evaluate whether a single injected fact can transfer successfully through multi-hop reasoning in language models. However, prior work has identified various issues in the original dataset, including logically invalid reasoning chains and incorrect entity mappings. We therefore use MQuAKE-Remastered, a cleaned and improved version of the benchmark. Specifically, we use the CF-3k split, employ cloze prompts for editing and Reliability evaluation, and use paraphrased question-form prompts to evaluate Generalization. Because the dataset does not provide Locality labels, we additionally construct a Locality set of 1,000 samples by selecting questions that do not overlap with those used for editing.

\paragraph{ZSRE}
ZSRE is a context-free QA dataset originally introduced for zero-shot relation extraction from Wikipedia text, but it has since become a standard benchmark for evaluating whether model editing methods can modify a model's implicit knowledge of specific relational tuples. We preprocess the ZSRE test set by removing duplicate items that may cause editing conflicts, resulting in a final evaluation set of 743 examples.

Table~\ref{tab:edit_dataset_examples} presents example instances from the MQuAKE-Remastered and ZSRE datasets. Table~\ref{tab:base_performance} reports the TF performance of LLaMA-3.1-8B-Instruct on MQuAKE and ZSRE before any knowledge editing is applied.
\begin{itemize}
\item \textbf{Target True:} the proportion of cases in which the base model correctly generates the original answer already encoded through pretraining.
\item \textbf{Target New:} the probability that the model generates the newly injected counterfactual knowledge by chance before editing.
\end{itemize}

\subsection{Preference Dataset}
To prevent degradation in generation quality under accumulated edits, GLIME computes a preference optimization loss using edit-independent random contexts $u$ sampled separately from the edit trigger $x_t$. 
For this purpose, we use OpenHermesPreferences, a large-scale synthetic preference dataset for alignment. Each instance consists of a user query ($u$), along with a preferred response ($y^+$) and a less preferred response ($y^-$), constructed by ranking generations from multiple LLMs with PairRM~\cite{llm-blender-2023}. Table~\ref{tab:pref_dataset_examples} shows example preference pairs used in training.

\section{Experimental Setting Details}
\label{sec:ap_setup}
Additional experimental details and hyperparameter settings not covered in the main text are provided below.

\subsection{Algorithms of GLIME}
For the pseudocode of GLIME, please refer to Algorithm~\ref{alg:glime}.

\begin{algorithm}[hbt!]
\caption{GLIME}
\label{alg:glime}
\small
\begin{algorithmic}[1]
\Require Initial model $\theta_0$; edit stream $\{e_t=(x_t, y_t^{\mathrm{new}})\}_{t=1}^{T}$; preference dataset $\mathcal{D}_{\mathrm{pref}}$; editable MLP layers $\mathcal{S}$; replay buffer size $M$; replay weight $\lambda_{\mathrm{rep}}$; basis size $k$; epochs $E$; learning rate $\eta$
\State Initialize replay buffer $\mathcal{B}\leftarrow \emptyset$
\ForAll{$\ell \in \mathcal{S}$}
    \State Initialize orthogonal basis $B^\ell \leftarrow [\ ]$
\EndFor
\State $\theta \leftarrow \theta_0$

\For{$t=1$ to $T$}
    \For{$j=1$ to $E$}
        \State Sample a preference pair $(u, y^+, y^-)\sim \mathcal{D}_{\mathrm{pref}}$
        \State Sample a replay mini-batch $\mathcal{R}\sim \mathcal{B}$
        \State Compute
        \[
\begin{aligned}
\mathcal{L}
&=
\mathcal{L}_{\mathrm{edit}}(\theta; e_t)
+
\mathcal{L}_{\mathrm{pref}}(\theta; u, y^+, y^-) \\
&\quad+
\lambda_{\mathrm{rep}}
\frac{1}{|\mathcal{R}|}
\sum_{e\in\mathcal{R}}
\mathcal{L}_{\mathrm{edit}}(\theta; e)
\end{aligned}
\]
        \ForAll{$\ell \in \mathcal{S}$}
            \State Compute layer gradient $G^\ell \leftarrow \nabla_{W^\ell}\mathcal{L}$
            \If{$B^\ell \neq [\ ]$}
                \State Project gradient:
                \[
                \tilde{G}^\ell \leftarrow G^\ell \left(I - B^\ell (B^\ell)^\top\right)
                \]
            \Else
                \State $\tilde{G}^\ell \leftarrow G^\ell$
            \EndIf
            \State Update parameters:
            \[
            W^\ell \leftarrow W^\ell - \eta \tilde{G}^\ell
            \]
            \State Update basis:
            \[
            B^\ell \leftarrow \textsc{UpdateBasis}(B^\ell, \tilde{G}^\ell, k)
            \]
        \EndFor
    \EndFor
    \State Update replay buffer:
    \[
    \mathcal{B} \leftarrow \textsc{UpdateBuffer}(\mathcal{B}, e_t, M)
    \]
\EndFor

\State \Return $\theta$
\end{algorithmic}
\end{algorithm}

\subsection{Prompt Templates}
We deliberately differentiate the prompt formats used in the editing and evaluation phases. Table~\ref{tab:prompt_templates} shows example prompt templates used in the editing and evaluation phases. 
\begin{itemize}
    \item \textbf{Editing Phase:} To directly inject factual knowledge into model parameters, we compute $\mathcal{L}_{\text{edit}}$ using a context-free cloze-style prompt, thereby avoiding unnecessary formatting bias introduced by system prompts or chat templates.
    
    \item \textbf{Evaluation Phase:} For AD-based evaluation, we generate text using an explicit QA prompt together with the official chat template for each model, so as to better reflect the real usage setting of instruction-tuned language models.
\end{itemize}

\subsection{Decoding Strategy}
For AD-based evaluation, all generations are produced using greedy decoding to encourage consistent generation of factual knowledge and control randomness. In addition, given the short-form nature of knowledge editing evaluation, we set max new tokens to 16 to prevent unnecessarily long generations.

\subsection{Hyperparameters}
We process all knowledge updates sequentially with a batch size of 1. We employ the AdamW optimizer for parameter updates, with the weight decay set to 0. The specific hyperparameter configurations are as follows:

\begin{itemize}\item \textbf{Base Hyperparameters:} The learning rate is fixed at $5 \times 10^{-5}$, and the number of training epochs per individual edit request is set to 3.\item \textbf{Replay Buffer:} For each update, $k=3$ samples are randomly retrieved from the replay buffer. To prevent overfitting to the current edit, the loss weight for the replay objective, $\lambda_{rep}$, is set to 0.1.\item \textbf{Gradient Projection:} To operate within practical memory constraints, we maintain a maximum of $r=4$ basis vectors for the orthogonal gradient projection. When this capacity is exceeded, the basis is updated online using a FIFO (First-In-First-Out) strategy, removing the oldest direction vectors to accommodate new ones.\end{itemize}

\section{More Analysis}
\label{sec:ap_analysis}

\begin{figure*}[t]
\centering 
\includegraphics[width=\textwidth]{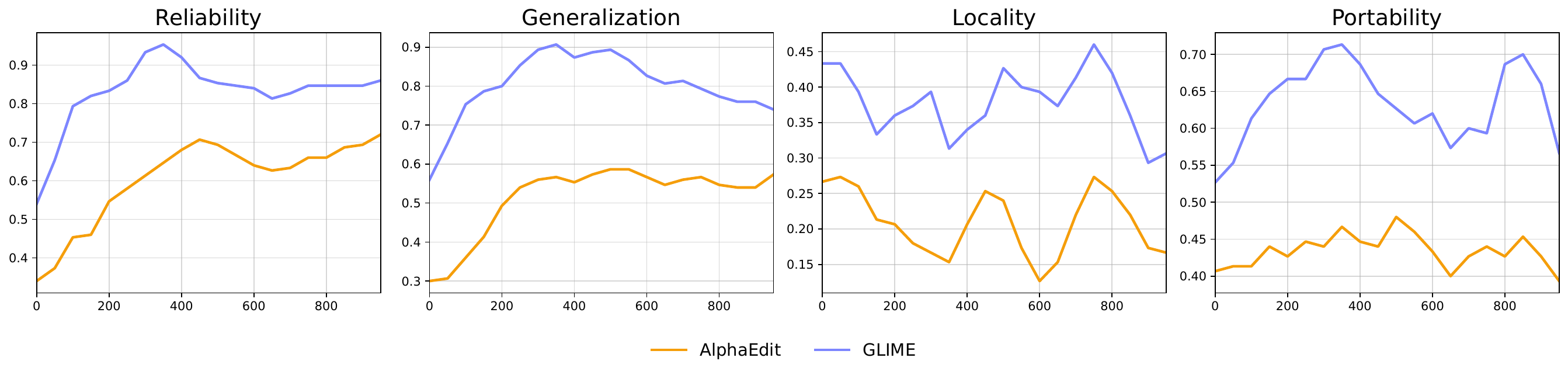}
\caption{Comparison of model outputs on each sample after 1,000 edits.}
\label{fig:seq_1000} 
\end{figure*}

\subsection{Knowledge Retention}
\label{subsec:retention_analysis}
Figure~\ref{fig:seq_1000} reports per-sample performance on each edit instance $e_i$, from 1 to 1000 edits, using the final model after all 1,000 edits have been completed. On Reliability and Generalization, AlphaEdit achieves relatively strong performance on recently injected knowledge (edits 800--1000), but exhibits a typical forgetting pattern in which performance drops sharply on older edits (edits 0--200). GLIME shows a similar trend, but maintains consistently higher edit success across the entire sequence. The gap between the two methods becomes even more pronounced on Portability. AlphaEdit remains around 0.4 across nearly all edit instances, suggesting limited ability to transfer edited facts to new contexts. In contrast, GLIME achieves high Portability throughout the full sequence, demonstrating that even older edited facts remain effectively usable in downstream reasoning.

\subsection{Semantic Evaluation via LLM-as-a-Judge}
\label{subsec:llm_as_a_judge_analysis}

\begin{table}[t]
\centering
\small
\setlength{\tabcolsep}{3.2pt}
\resizebox{\linewidth}{!}{%
\begin{tabular}{l}
\toprule
You are a strict grader. \\
\\
Given: \\
- Question \\
- Gold target \\
- Predicted answer \\
\\
Return: \\
A if the predicted answer semantically matches the gold target. \\
B otherwise. \\
\\
Only output a single letter: "A" or "B". \\
\\
Question: \{question\} \\
Gold target: \{target\} \\
Predicted answer: \{predicted\_answer\} \\
\bottomrule
\end{tabular}
}
\caption{LLM-as-Judge prompt template.}
\label{tab:llm_prompt}
\end{table}

\begin{table*}[t]
\centering
\small
\renewcommand{\arraystretch}{1.2}
\setlength{\tabcolsep}{4.5pt}
\begin{tabular}{l|ccc| ccc| ccc}
\toprule
 & \multicolumn{3}{c|}{\textbf{Reliability}} & \multicolumn{3}{c|}{\textbf{Generalization}} & \multicolumn{3}{c}{\textbf{Locality}} \\
\cmidrule(lr){2-4} \cmidrule(lr){5-7} \cmidrule(lr){8-10}
\textbf{Method} & \textbf{TF} & \textbf{AD} & \textbf{LLM} & \textbf{TF} & \textbf{AD} & \textbf{LLM} & \textbf{TF} & \textbf{AD} & \textbf{LLM} \\
\midrule
AlphaEdit & 0.912 & 0.604 & 0.594 & 0.497 & 0.514 & 0.500 & 0.324 & 0.209 & 0.173 \\
\rowcolor{myrow}\textbf{GLIME} & \textbf{0.937} & \textbf{0.830} & \textbf{0.775} & \textbf{0.806} & \textbf{0.800} & \textbf{0.746} & \textbf{0.517} & \textbf{0.379} & \textbf{0.328} \\
\bottomrule
\end{tabular}
\caption{Comparative analysis across evaluation metrics, including LLM-as-a-Judge.}
\label{tab:llm_judge_analysis}
\end{table*}

AD-based evaluation is straightforward to implement and easy to automate, as it is based on whether the correct answer is included in the response. However, this approach may not fully capture semantic correctness, since the incidental inclusion of the answer string can still be counted as correct. To address this limitation, we additionally employ an LLM-as-a-Judge~\cite{gu2025surveyllmasajudge} evaluation for AD responses. We use the instruction prompt shown in Table~\ref{tab:llm_prompt} and employ GPT-4o~\cite{openai2023gpt4} as the judge model. Table~\ref{tab:llm_judge_analysis} reveals a clear difference in performance retention between the two methods across evaluation protocols. Under LLM-as-a-Judge evaluation, which assesses strict semantic correctness, AlphaEdit continues to struggle, whereas GLIME preserves relatively strong performance. These results indicate that the high AD scores of GLIME do not merely arise from malformed outputs that happen to contain the correct answer token, but instead reflect the generation of contextually appropriate and semantically correct responses.

\subsection{Sensitivity Analysis of Hyperparameters}
\label{subsec:hyperparameter_analysis}

To identify an effective configuration for GLIME, we conduct a sensitivity analysis over its key hyperparameters. Table~\ref{tab:hyperparam_sensitivity} shows how performance varies with changes in the number of training epochs, replay batch size, replay loss weight ($\lambda_{rep}$), orthogonal basis size ($k$), and random seed. The default setting is $\text{training epoch}=3$, $\text{batch size}=3$, $\lambda_{rep}=0.1$, $k=256$, and $\text{seed}=7$.

\begin{table}[h]
\centering
\small
\renewcommand{\arraystretch}{1.1}
\setlength{\tabcolsep}{5pt}
\resizebox{\linewidth}{!}{%
\begin{tabular}{l| cccc}
\toprule
 & \textbf{Reliability} & \textbf{Generalization} & \textbf{Locality} & \textbf{Portability} \\
\midrule
\rowcolor{myrow}\textbf{Default} & 0.830 & 0.800 & 0.379 & 0.633 \\
\midrule
\multicolumn{5}{l}{\textit{Training Epoch}} \\ \midrule
1 & 0.653 & 0.665 & 0.436 & 0.507 \\
2 & 0.770 & 0.744 & 0.350 & 0.555 \\
4 & 0.796 & 0.798 & 0.308 & 0.608 \\
5 & 0.829 & 0.811 & 0.288 & 0.631 \\
\midrule
\multicolumn{5}{l}{\textit{Replay Batch Size}} \\ \midrule
1 & 0.767 & 0.751 & 0.341 & 0.568 \\
5 & 0.830 & 0.817 & 0.367 & 0.518 \\
7 & 0.851 & 0.843 & 0.351 & 0.570 \\
\midrule
\multicolumn{5}{l}{\textit{Replay Loss Weight ($\lambda_{rep}$)}} \\ \midrule
0.05 & 0.743 & 0.681 & 0.096 & 0.182 \\
0.20 & 0.824 & 0.814 & 0.298 & 0.575 \\
\midrule
\multicolumn{5}{l}{\textit{Basis Size ($k$)}} \\ \midrule
32 & 0.614 & 0.530 & 0.077 & 0.216 \\
64 & 0.782 & 0.770 & 0.350 & 0.629 \\
128 & 0.826 & 0.809 & 0.332 & 0.619 \\
512 & 0.819 & 0.822 & 0.345 & 0.598 \\
\midrule
\multicolumn{5}{l}{\textit{Gradient Rank ($r$)}} \\ \midrule 2 & 0.832 & 0.818 & 0.425 & 0.534 \\
\midrule
\multicolumn{5}{l}{\textit{Seed}} \\ \midrule
42 & 0.834 & 0.795 & 0.400 & 0.579 \\
\bottomrule
\end{tabular}
}
\caption{Sensitivity analysis results of GLIME under different hyperparameter settings. }
\label{tab:hyperparam_sensitivity}
\end{table}

\begin{itemize}
    \item \textbf{Training Epochs:} As the number of training epochs increases from 1 to 5, accumulated parameter updates gradually improve target knowledge injection and reasoning-based transfer. However, this improvement comes with a clear trade-off: Locality, which reflects preservation of existing knowledge, declines from 0.436 to 0.288.
    \item \textbf{Replay Batch Size \& Loss Weight ($\lambda_{rep}$):} These results reflect the role of experience replay in preserving previously edited knowledge. As the replay batch size increases, edit success improves slightly, but at the cost of higher computational overhead. 
    In particular, as the replay loss weight increases, edit success improves, but Locality declines, suggesting a trade-off caused by overly strong retention of edited knowledge.
    \item \textbf{Basis Size ($k$):} The basis size determines how sufficiently important update directions from past edits can be captured and preserved. When the basis size is excessively small, such as 32, it fails to adequately represent the principal subspace of previous edits, weakening the interference suppression effect and causing a substantial overall performance drop. In contrast, when the basis size reaches around 256, the evaluation metrics begin to converge, indicating the most stable generalization performance.
    \item \textbf{Gradient Rank ($r$):} The gradient rank controls the number of directions retained for each gradient subspace. Reducing $r$ from 4 to 2 slightly improves Reliability, Generalization, and Locality, but decreases Portability from 0.633 to 0.534. This suggests that a smaller rank may insufficiently preserve important update directions from earlier edits, thereby weakening the protection of transferable edited knowledge. We therefore use $r=4$ as the default setting to maintain a better balance across editing metrics.
    \item \textbf{Seed:} To verify robustness to random initialization, we compare performance by changing the default seed from 7 to 42. The results show that Reliability and Generalization remain at similar levels with little difference, while Locality slightly improves and Portability decreases marginally. Although some variation is observed in individual metrics, the overall performance trend remains unchanged, indicating that GLIME is relatively stable with respect to changes in the random seed.
\end{itemize}

\begin{table}[t]
\centering
\small
\setlength{\tabcolsep}{4pt}
\resizebox{\linewidth}{!}{%
\begin{tabular}{l|ccccc}
\toprule
\textbf{Method}
& \textbf{Reliability} & \textbf{Generalization} & \textbf{Locality}
& \textbf{Portability$_{\mathrm{MC}}$} & \textbf{Portability$_{\mathrm{MR}}$} \\
\midrule
\rowcolor{myrow}
\textbf{Sketch}
& 0.830 & 0.800 & 0.379 & 0.633 & \textbf{0.184} \\
SVD
& \textbf{0.871} & \textbf{0.838} & \textbf{0.394} & \textbf{0.682} & 0.181 \\
\bottomrule
\end{tabular}
}
\caption{Comparison between the randomized sketch and exact SVD for the gradient-space constraint (GC).}
\label{tab:svd_gc}
\end{table}

\subsection{Comparison with Exact SVD}
To examine the trade-off introduced by the sketch approximation in GC, we replace the randomized sketch with an exact SVD basis while keeping all other settings unchanged. As shown in Table~\ref{tab:svd_gc}, using exact SVD improves most metrics, increasing the average score from 0.565 to 0.593. However, constructing the exact SVD basis requires approximately $6\times$ more computation time than our sketch-based approach. Since lifelong model editing requires repeated and efficient updates as new edits arrive, this additional computational cost becomes substantial over long edit sequences. We therefore adopt the randomized sketch in GLIME as a practical trade-off, retaining competitive editing performance while substantially reducing the computational overhead of GC.

\begin{table}[t]
\centering
\small
\setlength{\tabcolsep}{4pt}
\resizebox{\linewidth}{!}{%
\begin{tabular}{l|ccccc}
\toprule
\textbf{Method}
& \textbf{Reliability} & \textbf{Generalization} & \textbf{Locality}
& \textbf{Portability$_{\mathrm{MC}}$} & \textbf{Portability$_{\mathrm{MR}}$} \\
\midrule

AlphaEdit
& 0.604
& 0.514
& 0.209
& 0.434
& 0.079 \\

+ Replay
& 0.530
& 0.545
& 0.166
& 0.320 
& 0.085 \\

\bottomrule
\end{tabular}
}
\caption{
Effect of applying memory replay to AlphaEdit.
}
\label{tab:replay}
\end{table}

\subsection{Effect of Memory Replay}
To examine whether GLIME's gains primarily stem from memory replay, we also apply the same replay strategy to AlphaEdit. As shown in Table~\ref{tab:replay}, Replay slightly improves Generalization from 0.514 to 0.545, but degrades Reliability, Locality, and Portability$_{\mathrm{MC}}$, resulting in a lower overall average score. These results indicate that replay alone is insufficient for effective lifelong editing. Rather, the improvements of GLIME arise from the complementary interaction between replay-based knowledge retention and the other components of our framework.

\begin{figure}[t]
\centering 
\includegraphics[width=0.8\linewidth]{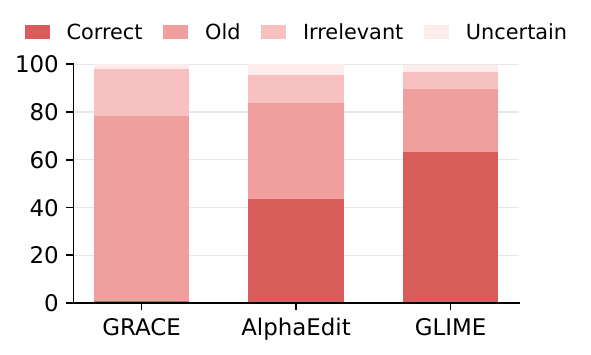}
\caption{Comparison of answers on the multiple-choice task after editing. \textit{Correct} denotes the proportion of selections corresponding to the edited knowledge, \textit{Old} denotes selections corresponding to the pre-edit knowledge, \textit{Irrelevant} denotes selections of unrelated answers, and \textit{Uncertain} denotes selections of the uncertainty option.}
\label{fig:mc} 
\end{figure}

\subsection{Multiple-Choice Answer Distribution}
Figure~\ref{fig:mc} compares the distribution of answers selected by each model on the multiple-choice task after editing. GRACE shows a strong tendency to rely on pre-edit knowledge, with most responses remaining in the \textit{Old} category, indicating limited transfer of edited facts to downstream decision-making tasks. AlphaEdit partially improves the \textit{Correct} ratio, but still retains a substantial proportion of \textit{Old} responses, suggesting that the use of edited knowledge is not yet stably established. In contrast, GLIME substantially reduces the proportion of \textit{Old} responses, showing that the model goes beyond producing the correct answer only for the edit prompt and instead applies the updated knowledge more consistently in decision-oriented queries. Moreover, the proportions of \textit{Irrelevant} and \textit{Uncertain} responses do not increase excessively, suggesting that GLIME also suppresses undesirable side effects such as random guessing and response uncertainty during editing.

\subsection{Loss Scale Analysis}
Since GLIME jointly optimizes multiple objectives, we examine whether differences in the magnitudes of their losses cause an imbalance during optimization. Table~\ref{tab:loss_scale} reports the average magnitude of each objective across training epochs. Although $\mathcal{L}_{edit}$ is substantially larger than $\mathcal{L}_{pref}$ and $\mathcal{L}_{replay}$ in the first epoch, it rapidly decreases as training proceeds and becomes comparable to the other objectives by the third epoch. This indicates that the initial scale discrepancy does not persist throughout optimization.

\begin{table}[t]
\centering
\small
\setlength{\tabcolsep}{8pt}
\begin{tabular}{c|ccc}
\toprule
\textbf{Epoch}
& $\mathcal{L}_{edit}$
& $\mathcal{L}_{pref}$
& $\mathcal{L}_{replay}$ \\
\midrule
1 & 2.8943 & 0.6979 & 0.5638 \\
2 & 1.1538 & 0.5074 & 0.5545 \\
3 & 0.5844 & 0.4319 & 0.5439 \\
\bottomrule
\end{tabular}
\caption{Average magnitude of each training objective across epochs.}
\label{tab:loss_scale}
\end{table}

To further examine whether explicitly compensating for the larger initial magnitude of $\mathcal{L}_{edit}$ is beneficial, we reduce its weight from the default value of 1.0 to 0.5. This results in lower Reliability (0.830 to 0.813), Generalization (0.800 to 0.777), and Portability (0.633 to 0.570), while Locality increases from 0.379 to 0.535. The overall average score also decreases from 0.565 to 0.558. These results suggest that explicitly down-weighting $\mathcal{L}_{edit}$ does not improve the overall balance among the objectives, supporting our default weighting scheme.

\section{Qualitative Analysis}
\label{sec:qualitative_analysis}
To provide an intuitive illustration of how GLIME preserves generation quality under lifelong knowledge editing, we present decoding outputs of the language model in Table~\ref{tab:qualitative_case_study}. The example is taken after 1,000 accumulated edits, including an edit that changes the nationality of the target person (Frank R. Strayer) from \textit{American} to \textit{Canadian}. The table shows the model's autonomous generations for the prompts corresponding to each evaluation metric.

\begin{table*}[t]
\centering
\small
\renewcommand{\arraystretch}{1.25}
\begin{tabular}{p{2.2cm} p{13cm}}
\toprule
\textbf{Method} & \textbf{Predicted Answers} \\
\midrule

\multicolumn{2}{l}{\textbf{[Reliability]}} \\
\multicolumn{2}{l}{\textbf{Prompt:} "Frank R. Strayer is a citizen of" \quad \textbf{Target:} Canada} \\
\midrule
FT-L & ://:// of://://:// of:// of \textcolor{red}{\faTimes} \textit{(Model Collapse)} \\
R-ROME & INTERRUPTION INTERRUPTION Ras INTERRUPTIONdig... \textcolor{red}{\faTimes} \textit{(Model Collapse)} \\
GRACE & I do not have information on a person named Frank R. Strayer. \textcolor{red}{\faTimes} \textit{(Refusal)} \\
MEMIT & United United United United United United... \textcolor{red}{\faTimes} \textit{(Repetition)} \\
WISE & Canada Canada Canada Canada Canada Canada... \textcolor{red}{\faTimes} \textit{(Repetition)} \\
AlphaEdit & I don't have information on Frank R. Strayer. \textcolor{red}{\faTimes} \textit{(Refusal)} \\
\rowcolor{blue!5} \textbf{GLIME} & \textbf{Canada.} \textcolor{green}{\faCheck} \\
\midrule

\multicolumn{2}{l}{\textbf{[Generalization]}} \\
\multicolumn{2}{l}{\textbf{Prompt:} "What is the country of citizenship of Frank R. Strayer?" \quad \textbf{Target:} Canada} \\
\midrule
FT-L & :// of:// of://:// of:// of:// \textcolor{red}{\faTimes} \\
R-ROME & dyst pers**reeze** INTERRUPTION INTERRUPTION.scalablytyped... \textcolor{red}{\faTimes} \\
GRACE & I do not have information on the country of citizenship of Frank R. Strayer \textcolor{red}{\faTimes} \\
MEMIT & United United United United United United... \textcolor{red}{\faTimes} \\
WISE & Canada Canada Canada Canada Canada Canada... \textcolor{red}{\faTimes} \\
AlphaEdit & I don't have information on Frank R. Strayer. \textcolor{red}{\faTimes} \\
\rowcolor{blue!5} \textbf{GLIME} & \textbf{Frank R. Strayer is a citizen of Canada.} \textcolor{green}{\faCheck} \\
\midrule

\multicolumn{2}{l}{\textbf{[Locality]}} \\
\multicolumn{2}{l}{\textbf{Prompt:} "Who founded Tudor dynasty?" \quad \textbf{Target:} Henry VII \textit{(Pre-trained knowledge)}} \\
\midrule
FT-L & ://://://:// of:// of:// of \textcolor{red}{\faTimes} \\
R-ROME & .scalablytyped.scalablytyped.scalablytyped.scalablytyped... \textcolor{red}{\faTimes} \\
GRACE & The Tudor dynasty was founded by Henry Tudor, also known as Henry VII \textcolor{green}{\faCheck} \\
MEMIT & United United United United United United... \textcolor{red}{\faTimes} \\
WISE & Canada Canada Canada Canada Canada Canada... \textcolor{red}{\faTimes} \textit{(Knowledge Bleeding)} \\
AlphaEdit & The founder of the Tudor dynasty was John of Ga. \textcolor{red}{\faTimes} \textit{(Hallucination)} \\
\rowcolor{blue!5} \textbf{GLIME} & \textbf{The Tudor dynasty was founded by Henry Tudor, who was born in the...} \textcolor{green}{\faCheck} \\
\midrule

\multicolumn{2}{l}{\textbf{[Multi-hop Reasoning / Portability]}} \\
\multicolumn{2}{l}{\textbf{Prompt:} "Who is the current head of the government in the country of citizenship of Frank R. Strayer?"} \\
\multicolumn{2}{l}{\textbf{Target:} Justin Trudeau \textit{(Canada's head of government)}} \\
\midrule
FT-L & of:// of://://:// of://:// of \textcolor{red}{\faTimes} \\
R-ROME & grisione.scalablytyped.scalablytyped.scalablytyped... \textcolor{red}{\faTimes} \\
GRACE & I am unable to verify who the current head of government is for Frank R. \textcolor{red}{\faTimes} \\
MEMIT & United United United United United United... \textcolor{red}{\faTimes} \\
WISE & Canada Canada Canada Canada Canada Canada... \textcolor{red}{\faTimes} \\
AlphaEdit & I don't have information on the current head of the government in the country of... \textcolor{red}{\faTimes} \\
\rowcolor{blue!5} \textbf{GLIME} & \textbf{The current head of government in the country of Canada is Justin Trudeau, who is...} \textcolor{green}{\faCheck} \\

\bottomrule
\end{tabular}
\caption{Qualitative evaluation of language model generations after accumulated edits. }
\label{tab:qualitative_case_study}
\end{table*}

\begin{itemize}
    \item \textbf{Model Collapse (FT-L, R-ROME, MEMIT):} FT-L and R-ROME exhibit complete collapse of decoding ability, generating meaningless special symbols (e.g., ``://'') or corrupted text (e.g., ``scalablytyped''). MEMIT also suffers severe collapse under exposure bias, repeatedly generating the same word (``United'') without termination.
    \item \textbf{Target Bleeding (WISE):} Although WISE avoids complete collapse through its dual-memory architecture, it excessively amplifies the probability of the target token (``Canada''), causing knowledge bleeding in which the model outputs ``Canada'' even for prompts about unrelated historical facts.
    \item \textbf{Refusal \& Hallucination (GRACE, AlphaEdit):} GRACE and AlphaEdit preserve fluency relatively well, but fail to integrate the edited knowledge into the reasoning process and instead refuse to answer. In particular, AlphaEdit also exhibits hallucination on the Locality prompt by generating an irrelevant historical fact.
    \item \textbf{GLIME:} GLIME preserves language quality through behavior-level preference optimization. Beyond merely recalling the target answer, it integrates ``Canada'' into a complete and natural sentence, while also demonstrating robust performance on both multi-hop reasoning and unrelated factual knowledge.
\end{itemize}

\end{document}